# Robust Encoder-Decoder Learning Framework towards Offline Handwritten Mathematical Expression Recognition Based on Multi-Scale Deep Neural Network

Guangcun Shan[1*], Hongyu Wang[1] &Wei Liang[2,3]

[1]School of Instrumentation Science and Optoelectronics Engineering, Beihang University, Beijing 100191, China;
[2]Institute of Electronics, Chinese Academy of Sciences, Beijing 100190, China;
[3]University of Chinese Academy of Sciences, Beijing 100049, China;

* Corresponding author. Tel.: +86-10-82339603 or +852-34422825; fax: +852-34420538.

**Abstract** Offline handwritten mathematical expression recognition is a challenging task, because handwritten mathematical expressions mainly have two problems in the process of recognition. On one hand, it is how to correctly recognize different mathematical symbols. On the other hand, it is how to correctly recognize the two-dimensional structure existing in mathematical expressions. Inspired by recent work in deep learning, a new neural network model that combines a Multi-Scale convolutional neural network (CNN) with an Attention recurrent neural network (RNN) is proposed to identify two-dimensional handwritten mathematical expressions as one-dimensional LaTeX sequences. As a result, the model proposed in the present work has achieved a WER error of 25.715% and ExpRate of 28.216%.

## 1 Introduction

Mathematical expressions are now widely used in scientific research, finance, and statistics, and are closely related to our daily life. Each one of us will have to be exposed to a large number of mathematical expressions in everyday learning through different ways, which may include newspapers, journal documents, notes, and so on. Sometimes, especially for researchers, it is inevitable to record some mathematical expressions on computers. However, the mathematical expressions are usually complicated. If people input them one by one, it will take a lot of time. So, if computers can automatically identify these expressions, the research will become more convenient (like reading and writing scientific papers). Moreover, it is also helpful for educating because students can understand new mathematical expressions without teacher  if they can be identified. Therefore, how to let computers automatically recognize mathematical expressions has become a hot research issue. This problem has been studied for decades and first proposed by [1]. Then, many researchers  have tried to study this problem at different paces. Chan and Yeung[2] proposed a method that is used definite clause grammars, the Lavirotte and Pottier[3] proposed a model which is based on graph grammars. Yamamoto et al.[4] presented a system using Probabilistic Context-Free Grammars(PCFG), and MacLean and Labahn [5] developed an approach using relational grammars and fuzzy sets. The recognition of mathematical expressions can be divided into two categories by the recognition objects: print mathematical expression recognition and handwritten mathematical expression recognition. A print mathematical expression is shown in Figure 1 and a handwritten mathematical expression recognition is shown in Figure 2.

For the recognition of print mathematical expressions, it is easy to identify because of its neat mathematical structure, clear writing, and clear spatial location. However, for the handwritten mathematical expressions, the recognition process may have a series of problems such as blurred writing and unclear spatial positions due to different writing habits of the writer. At present, researchers have achieved some results in the field of mathematical expression recognition. In the recognition of print mathematical expressions, Kumar et al. [6] proposed an identification method consisting of three stages: symbol generation, structure analysis and code generation. Yoo et al. [7] proposed an improved recursive projection contour cutting method for character segmentation of mathematical expressions. Amit et al. [8] proposed a new method using multiple structure recognition algorithms. In recognition of handwritten mathematical expressions, the identification method proposed by Hu et al. [9] is based on the segmentation of handwritten strokes. Zhang et al. [10] proposed a model using Watch, Attend, and Parse to perform image recognition that includes mathematical expressions.

$$e^{\theta} = \sum_{n=0}^{\infty} \frac{\theta^n}{n!}$$

**Figure 1** A print mathematical expression

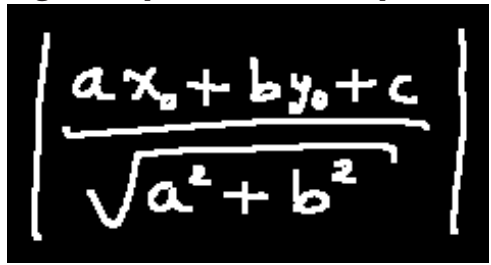

**Figure 2.** A handwritten mathematical expression

Compared to the traditional text recognition, there remains still some difficulties in the recognition of mathematical expressions as a challenging research problem. The main reason is that mathematical expressions have a large number of mathematical symbols and a complex two-dimensional structure. First of all, the character coverage of mathematical expressions includes numbers, operators, English letters, Roman letters, commas, dots, and other types. Secondly, some symbols are not shown in the expressions and these symbols need to be recognized by the spatial position of the mathematical expressions. For example, "X⁶" means the sixth power of "X", but the operator "^" is not directly written in this mathematical expression. Finally, some of the characters in the mathematical expression are very similar but the meaning is completely different. For example, the English letter "O" and the number "0". Therefore, due to the existence of these problems, handwritten mathematical expression recognition has much room for development in terms of the accuracy and range of recognition. This paper is organized as follows. In Section 2, the basic encoder-decoder framework and algorithm of our deep neural network model is presented. After that, the data training and testing of our neural network will be elaborated in Section 3. Then, in Section 4, the experimental results and the visualiza-tion of datasets are illustrated and discussed in details. Finally, conclusions are presented in Section 5.

## 2   Framework and model

### 2.1  Encoder-Decoder Framework

The Encoder-Decoder framework is a very common framework in deep learning. It consists of three parts, which are Encoder, Decoder and Context Vector generated by the Encoder. The structure of Encoder-Decoder framework is shown in Figure 3 and this framework is flexible enough because different neural networks can be selected as encoder and decoder for different tasks. Generally, convolutional neural network (CNN) is used for image data, and recurrent neural network (RNN) is used for sequence data. So, in this paper, we use CNN as our encoder because the input are pictures and we use RNN as our decoder because the output are LaTex Sequences.

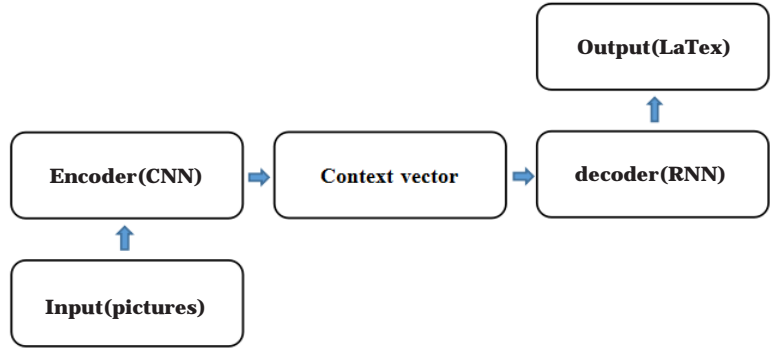

**Figure 3** The structure of Encoder-Decoder Framework

As shown in Figure 3, input and output can be expressed as equation (1) and equation (2). The $x_i$ means the i-th picture in input data and the $y_{i1}$, $y_{i2}$..... means the output(LaTex Sequence) of i-th picture in our model.

$$X = x_i \tag{1}$$

$$Y = (y_{i1}, y_{i2}, y_{i3}, ..., y_{im}) \tag{2}$$

The role of the encoder is to extract features of the input picture. Therefore, the output of the encoder can be expressed by:

$$Context = F(x_i) \tag{3}$$

When we have a context vector, the decoder will decode this context vector and generate the output. So, the output of the decoder can be expressed by:

$$y_i = F^{-1}(Context, y_1, y_2, y_3, ..., y_{i-1}) \tag{4}$$

In equation (4), $y_1$ is the decoder's initial value, which is usually given by the user. The encoding-decoding framework is currently widely used in deep learning such as machine translation and picture annotation. For example, Dzmitry Bahdanau et al. [11] proposed a machine translation model based on this framework which encoder and decoder are both recurrent neural network. Andrej Karpathy et al. [12] proposed a picture annotation algorithm based on this framework which encoder is convolutional neural network and decoder is recurrent neural network.

## 2.2 Encoder--Multi-Scale Convolutional Neural Network

In this paper, we used a convolutional neural network called DenseNet as encoder and improved it. This network is proposed by Gao Huang in 2017[13]. We designed three Dense Blocks and plused a Multi-Scale structure for better extracting features of input picture. The Multi-Scale convolutional neural network used in this paper is shown in Figure 4.

In Figure 4, the part of DenseNet model has three Dense Blocks. Before each Dense Block, there are convolutional layer and pooling layer. The first convolutional layer utilizes a square convolution kernel with a size of $(7 \times 7)$, and the convolution stride size is fixed to $(2 \times 2)$. Then, the size of the convolution filters and stride are both $(1 \times 1)$ in the next convolutional layers. Moreover, the max-pooling layer and the average-pooling layer are both carried out on a $(2 \times 2)$ kernel, with a stride fixed to $(2 \times 2)$. Then, in the part of Multi-Scale model, we designed a $(1 \times 1)$ convolutional layers after the conv layer2, conv layer3 and feature maps and the role of these $1 \times 1$ conv layers are reducing the dimension of features for less computation. Next, we can have the feature maps C1, C2, C3 shown in Figure 4, and we will add C1 and C2 to a new C2, and add this new C2 to C3 for a new C3. Here, we must have a trouble as these feature maps are different size. Therefore, in this paper, we use a $2 \times$ upsample for C1 and C2 because in Densenet, after many convolution layers, the size of C3 is two times that of C2 and the size of C2 is two times that of C1. Finally, the C1, C2, C3 are our encoder's output.

The reason that why we use this structure is that we believe the lower feature maps will have more location information. On the contrary, though the higher feature maps will have more advanced features, they will loss many basic information as they are extracted after more convolution layers[14].

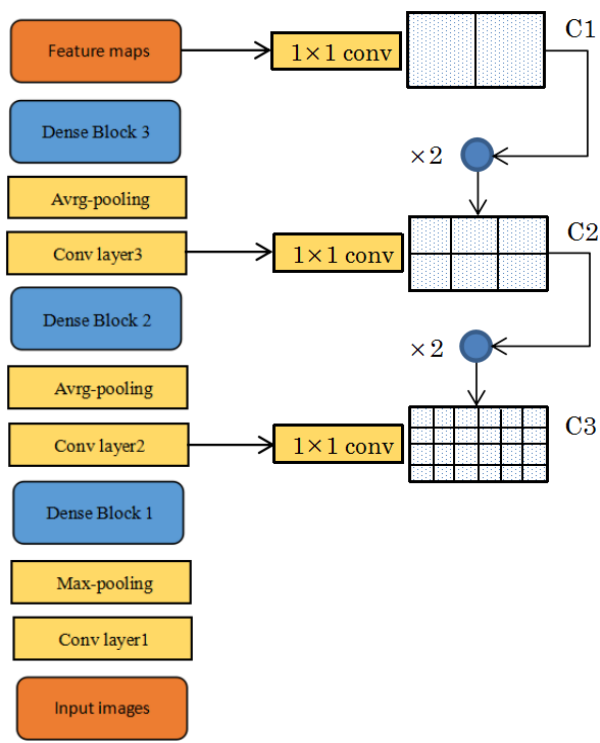

**Figure 4** The structure of DenseNet used in my model

## 2.3 Decoder--Recurrent Neural Network

### 2.3.1 Recurrent neural network parameter initialization

The decoder we used here is a recurrent neural network. In detail, the GRU model[15] is used in this work for the present design of the RNN. In the experiment, the first input of the recurrent neural network is defined as <sos> (start of string), indicating the beginning of the sequence. Meanwhile, the value of the first hidden layer in the recurrent neural network can be calculated by:

$$\overline{a} = \frac{1}{L}\sum_{i=1}^{L} a_i \tag{5}$$

$$s_0 = \tanh(W_{init}\overline{a}) \tag{6}$$

Among them, a is the advanced feature maps extracted by the convolutional neural network, W is a random initialization matrix, $s_0$ is the first hidden layer of the recurrent neural network.

### 2.3.2 Attention-based RNN model

Although the GRU model is used, the effect is still not ideal for complex sequences such as mathematical expressions. Therefore, the attention model is adopted in the present RNN. The attention model is a method that can greatly improve the performance of the RNN. Its main design ideas are described in the following.

For a long sequence like equation (7), the output that we want to get is equation (8). Obviously, especially for sequences that have a time-series relationship, when our output is $b_1$, the first part of the input sequence is mainly focused on. When our output is $b_m$, we hope that the end of the input sequence will then be focused on. However, when the traditional recurrent neural network outputs each result, the weight of each part in the input sequence is same.

$$A = (a_1, a_2, a_3, ..., a_n) \tag{7}$$

$$B = (b_1, b_2, b_3, ..., b_m) \tag{8}$$

For example, if our output of the recurrent neural network is equation (4), we can expand this output step-by-step. Then we can get the first few steps:

$$y_2 = F^{-1}(Context, y_1) \tag{9}$$

$$y_3 = F^{-1}(Context, y_1, y_2) \tag{10}$$

$$y_4 = F^{-1}(Context, y_1, y_2, y_3) \tag{11}$$

It can be found that for each output of y, the context vector computed by the RNN is same. However, after we add the attention model, the output of equations (9), (10), (11) will be given by:

$$y_2 = F^{-1}(Context_1, y_1) \tag{12}$$

$$y_3 = F^{-1}(Context_2, y_1, y_2) \tag{13}$$

$$y_4 = F^{-1}(Context_3, y_1, y_2, y_3) \tag{14}$$

After adding the attention model, the context in the output of each step will be changed according to the parts that need to attention. In our model, as the outputs of convolutional neural network are three three-dimensional matrixs and each one of them can be expressed as C×H×W, the inputs in our recurrent neural network can be shown in the following:

$$a = \{a_1, a_2, a_3, ..., a_L\}, a_i \in R^C \tag{15}$$

In equation (15), L=H×W. Thus, each $a_i$ is a C dimensional vector. Then, we define a function $f_{att}$ shonwen in equation (16) and (17) as our attention function. So, the RNN with attention in our model can be calculated by the function $f_{att}$.

$$e_{ti} = v_a^T \tanh(W_a h_{t-1} + U_a a_i) \tag{16}$$

$$\alpha_{ti} = \frac{\exp(e_{ti})}{\sum_{k=1}^{L} \exp(e_{tk})} \tag{17}$$

In equation (16), $h_{t-1}$ is the value of previous hidden layer in the GRU model, and the initial value of h is calculated by equation (6). $a_i$ is the i-th vector in equation (15) $v_a$ is a random initialization matrix and is updated continuously following the training of the network. Its main role is to adjust the dimensions of eti. In equation (17), $\alpha_{ti}$ is the value of next attention model.

### 2.3.3 Coverage model

For the recurrent neural network with attention model, its performance has been greatly improved compared to the recurrent neural network without attention model. But there is still a problem in attention model which is lack of coverage. Coverage means the parts of input images whether have been translated. If we do not add a coverage model, the computer may repeatedly translate one part. Coverage model need us to sum all past value of attention model. So, after including the coverage model, the function $f_{att}$ will be computed as follows and we define it as $f_{att\_cov}$:

$$\beta_t = \sum_{l}^{t-1} \alpha_l \tag{18}$$

$$F = Q * \beta_t \tag{19}$$

$$e_{ti} = v_a^T \tanh(W_a h_{t-1} + U_a a_i + U_f f_i) \tag{20}$$

$$\alpha_{ti} = \frac{\exp(e_{ti})}{\sum_{k=1}^{L} \exp(e_{tk})} \tag{21}$$

In equation (18), $\alpha$ is the value calculated by the attention model in the previous step, and its initial value is 0. In equation (19), Q is a random initialization matrix and is updated continuously following the training of the network. The main role of Q is also to adjust the dimensions. F is the value of coverage model, and $f_i$ is the i-th vector in F. Then, the attention value can be calculated by using the equation (20) and (21).

### 2.3.4 Output

According to our output from encoder is multi-scale, the context vector of each scale will be computed using the following equation:

$$c_t = \sum_{i}^{L} \alpha_{ti} a_i \tag{22}$$

In equation (22), $c_{c1}$ will use $f_{att\_cov\_c1}$ to calculate $\alpha_{c1}$, and the same with $c_{c2}$, $c_{c3}$. Then, we will combine them as $c_t$ by just using cat function. After that, we can compute the next hidden layer as follows:

$$h_t = GRU(x_{t-1}, h_{t-1}, c_t) \tag{23}$$

 In equation (23), $h_{t-1}$ is the value of previous hidden layer and $x_{t-1}$ is the value of previous input which initial value is <sos>. According to equation (23), we can get the next hidden layer $h_t$. Finally, our output from the model can be computed:

$$y_t = g(W_0(Ey_{t-1} + W_h h_t + W_c c_t)) \tag{24}$$

In equation (24), g is the softmax function. $W_0$, $W_h$, $W_c$, and E are random initialization matrixs that can be continuously updated following neural network training.

## 3   Training and Testing of Neural Network Model

### 3.1  Dataset

The dataset used in this paper is CROHME 2014 (now available at CROHME website) and it is the largest dataset in handwritten mathematical expression. It has a train dataset of 8836 math expressions and a test dataset of 986 math expressions. Meanwhile, there are 110 different math symbols including numbers, almost all common operators and two start and stop symbols <sos> , <eol>. After that, the size of pictures in CROHME 2014 is very uneven, which will make the recognition very difficult. In train and test dataset, the biggest picture size is almost 400,000 pixels and the smallest  is only 1,400 pixels.

### 3.2  Training and Testing

In deep learning, the main target of training is to continuously update the neural network's parameters for reducing the loss in prediction and truth. In experiment, we choose the Adam as our optimizer and CrossEntropy as our loss. The initial learning rate is 0.00010. After complete our training in training dataset, we will test our model in testing dataset and the model which can get the best result is our best model.

### 3.2.1  WER Loss

In this paper, we use the WER loss to evaluate the prediction performance of neural networks. WER loss is the ratio of edit distance between the prediction and the truth to the length of the truth. Edit distance is the smallest edit operation from one string to another. Allowed editing operations include: insert a character, delete a character, replace a character. An example of the edit distance is shown in Table 1.

**Table1** An Example of Edit Distance

| Edit operation | Example |
| --- | --- |
| Insert | Prediction：ab=1 Truth：a+b=1 |
| Delete | Prediction：a++b=1 Truth：a+b=1 |
| Replace | Prediction：a+c=1 Truth：a+b=1 |

If our prediction sequence is A, the real sequence is B. From A to B, a minimum of $W_1$ insertions, $W_2$ deletions, and $W_3$ replacements are required. Also know that the length of sequence B is $W_4$, then the WER loss of this prediction is:

$$WER = \frac{W_1 + W_2 + W_3}{W_4} \tag{25}$$

According to the WER loss, we can evaluate the quality of a model.

## 3.3 Results

The experiments in this paper were all done on NVIDIA Tesla K80. The CUDA version used in the experiment was 8.0, the cuDNN version was 7.0, the Python version was 3.6, and the Pytorch version was 0.3.

### 3.3.1 The Result in Different Learning Rate

In order to find the most suitable learning rate, this paper tested four cases with a learning rate of 0.00010-0.00040. The result is shown in Figure 5 and Figure 6.

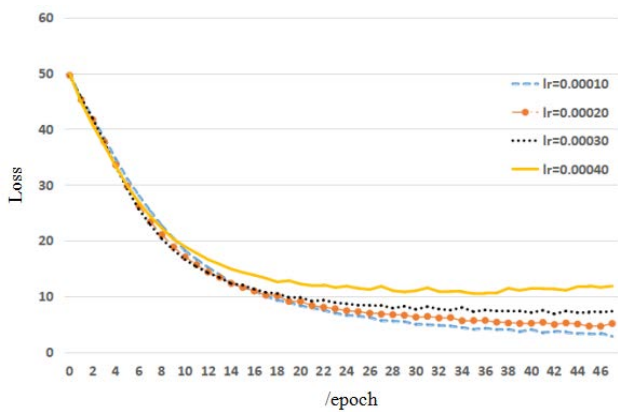

**Figure 5** The loss in training dataset I

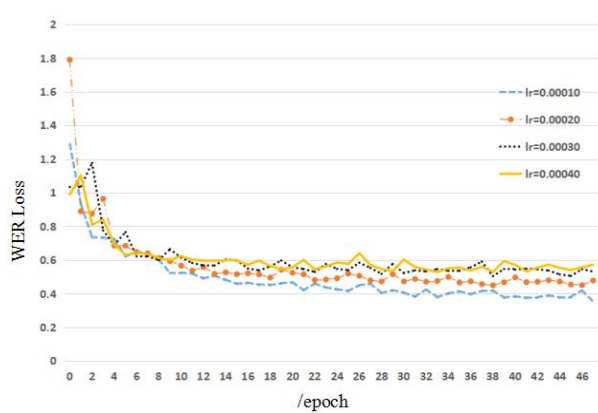

**Figure 6** The WER loss in testing dataset I

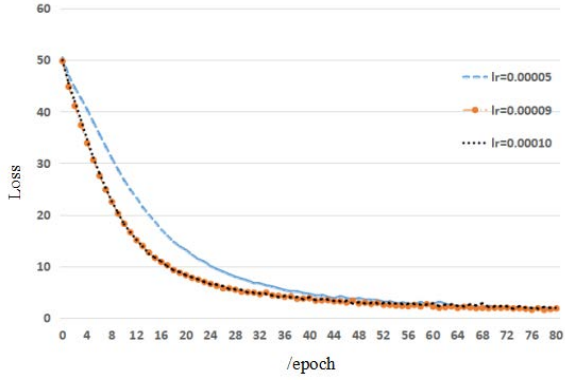
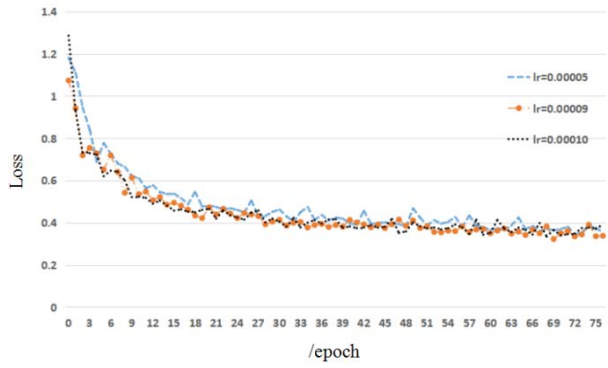

**Figure 7** The loss in training datase II      **Figure 8** The WER loss in testing dataset II

It can be seen that a learning rate of 0.00010 can achieve better results than other learning rates. Thus, this paper compares other three groups of learning rates, which are 0.00010, 0.00009, and 0.00005, respectively. From Figure 7 and Figure 8, we can see that when the learning rates are 0.00010 and 0.00009, the WER loss have better results than the learning rate is 0.00005. As a result, these models can be tested in testing program, and the results are shown in Table 2.

In Table 2, the ExpRate is the accuracy in testing dataset and we can also see that when the learning rate is 0.00009, the result is better than others.

**Table 2** The result of different learning rate in testing program

| Learning rate | WER Loss | ExpRate |
|---------------|----------|---------|
| 0.00009 | **28.619%** | 21.730% |
| 0.00010 | 29.961% | 21.730% |

### 3.3.2 Teacher-forcing

In order to further improve the ability of our model, this paper uses a method called Teacher-forcing.Teacher-forcing is specifically applied to training recurrent neural networks. The basic idea of this method is to use the truth values instead of the predictive values as the next input of recurrent neural network. If we do not use the Teacher-forcing method, then each step of the input to the recurrent neural network will be the predicted values, as shown in Figure 9 (a). If the Teacher-forcing method is used, the input for each step of the recurrent neural network will be the truth values, as shown in Figure 9 (b).

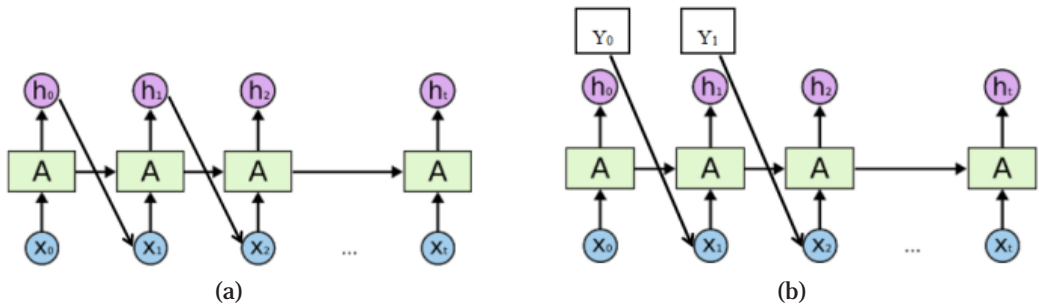

**Figure 9** The Teacher-forcing model, (a) is a RNN model without Teacher-forcing, (b) is a RNN model with Teacher-forcing

Using Teacher-forcing requires setting a replacement rate, which is the probability that the true value replaces the predicted value. This paper currently only tests the replacement rate of 20%. After using Teacher-forcing, the loss on the training dataset and the WER loss on the testing dataset are shown in Figure 10.

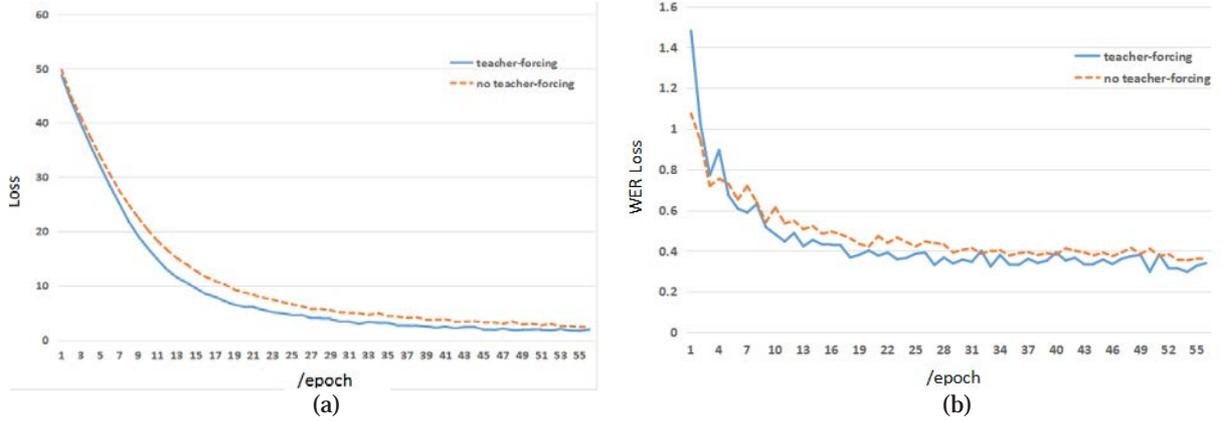

**Figure 10** (a) is the loss in Training dataset with or without using Teacher-forcing (b) is the WER loss in Testing dataset with or without using Teacher-forcing

Obviously, model that use Teacher-forcing can achieve better results than that do not use Teacher-forcing, either from the training dataset or the testing dataset. After verification on the testing program, the results shown in Table 3 can be obtained.

**Table 3.** The result of testing program in different learning rate

| Learning Rate | Teacher-forcing | | No Teacher-forcing | |
|---|---|---|---|---|
| | WER Loss | ExpRate | WER Loss | ExpRate |
| 0.00009 | **25.715%** | **28.216%** | 28.619% | 21.730% |
| 0.00010 | 28.089% | 26.162% | 29.961% | 21.730% |

As can be seen from Table.3, the WER loss decreases by 2.904% and the ExpRate increases by 6.486% when the learning rate is 0.00009 with Teacher-forcing. Meanwhile, with a learning rate of 0.00010, the WER loss decreased by 1.872% and the ExpRate increased by 4.432%.

Therefore, using Teacher-forcing has a positive effect on neural network training. And we can also see that the best testing result for the experiments performed in this paper were obtained by using Teacher-forcing at a learning rate of 0.00009, with a WER error of 25.715% and an ExpRate of 28.216%.

Table 4 shows the comparison between our result and the results of other researchers. It can be seen that our model has achieved a better result than others. In Table 4, Tokyo Univ.(The Nakagawa Labora-tory of Tokyo University of Agriculture and Technology) designed a system with three modules and the algorithm is mainly based on Stochastic Context Free Grammar[16-18], but its model is complicated. Univ.Nantes proposed a system that can consider the invalid symbol segments for preventing errors from one step to another[16,19,20]. This is a good idea for long sequence but not enough for mathematical expressions. RIT used a method that considered strokes in time series including number of strokes, aspect ratio, covariance matrix of sample points and so on[16,21-23]. However, its attention to the overall structure is less.

**Table 4** Results of our model and others

| System | ExpRate |
|---|---|
| Tokyo Univ. | 19.97% |
| Univ. Nantes | 18.33% |
| Rochester Institute of Technology (RIT) | 14.31% |
| ours | **28.21%** |

In summary, our model with attention and coverage can not only focus on the whole structure, but also pay attention to small details. Therefore, we can get the best results.

### 3.3.3 The Experiments of Coverage model

In order to better show the effectiveness of coverage model, we compared the result of the model with and without coverage. The experiments were completed at a learning rate of 0.00009 and using the Teacher-forcing.

In Figure 11, we can see that the WER loss in testing dataset has dropped significantly by using coverage model. In addition, we also compared the forecast result from the model with using coverage or without using coverage.

From Figure 12 and Figure 13, we can see that after using the coverage, the model correctly predicted the results. But, the model repeatedly outputs a "/theta" if we do not use the coverage.

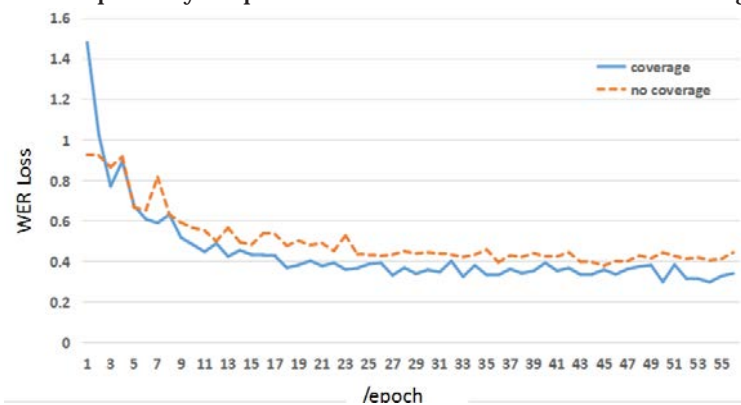

**Figure 11** The WER loss in Testing dataset with or without using Coverage model

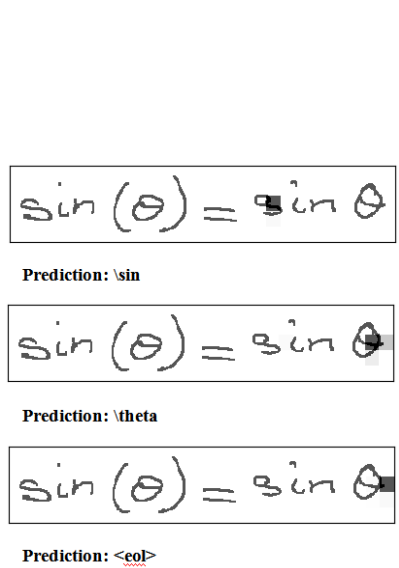

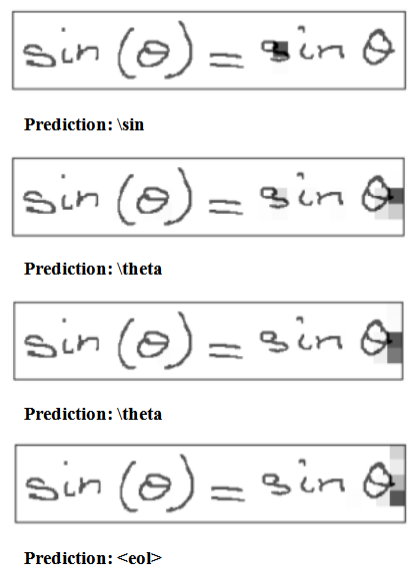

**Figure 12** The forecast result with coverage    **Figure 13** The forecast result without coverage

## 4 Visualization of Experiment Results

### 4.1 Introduction

In order to demonstrate the entire mathematical expression recognition process, the output of the model is visualized. The first convolution layer's output, and the output of each step in recurrent neural network were visualized.

## 4.2 Input Image

In this paper, we have 912 pictures in testing dataset and we randomly selected an input picture shown in Figure 14.

## 4.3 The first convolution layer's output

The first layer of the convolutional neural network used in this paper has 48 convolution kernels. Each convolution kernel corresponds to the advanced features of different parts of the original image. Therefore, the output of the first convolutional neural network has 48 different feature maps. Figure 15 randomly show four of the 48 features.

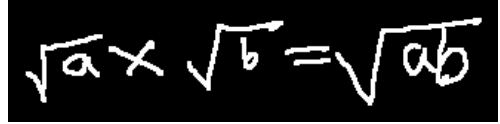

**Figure 14** An example of input image

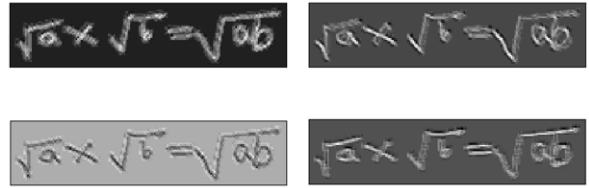

**Figure 15** The First Convolution Layer's Output

## 4.4 Output of each step in recurrent neural network

After the convolutional neural network extracts the advanced feature maps of pictures, this information will be passed to the recurrent neural network. The output of the recurrent neural network will continue until the network predicts the end <eol> (end of line). Figure 16 shows the step-by-step output of recurrent neural network.

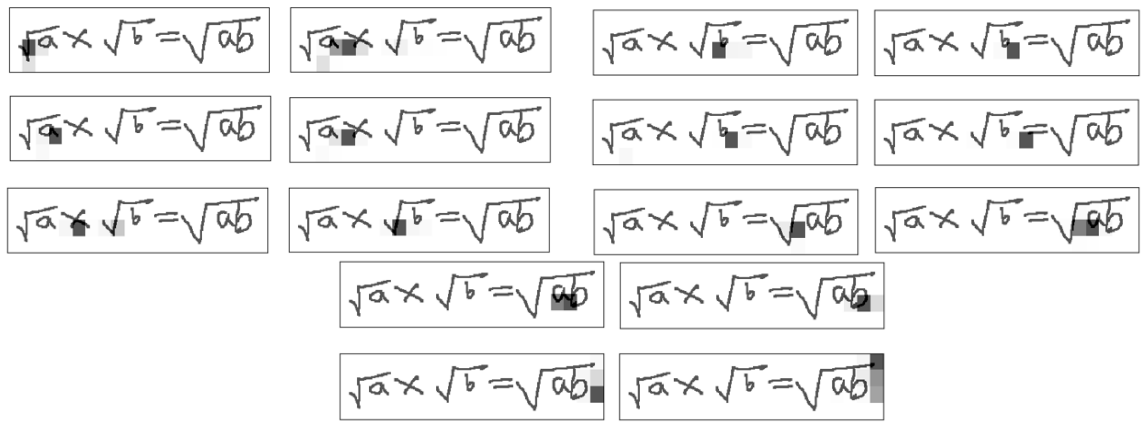

**Figure 16** The step-by-step output of recurrent neural network

In Figure 16, the black area is the result of attention model and represents the area of focus in this step. The prediction LaTeX result in Figure 16 is "\sqrt, {, a, }, \times, \sqrt, {, b, }, =, \sqrt, {, a, b, }, <eol>" which means $\sqrt{a} \times \sqrt{b} = \sqrt{ab}$ and the truth of this picture is also $\sqrt{a} \times \sqrt{b} = \sqrt{ab}$ . As a result, the model correctly predicts the result.

## 5   Conclusions

Based on the combination of a Multi-Scale convolutional neural network and an attention-based RNN, a new neural network model is performed to identify two-dimensional handwritten mathematical expressions as one-dimensional LaTeX sequences. When using CROHME 2014 dataset, we show the best experiment result that the WER loss and ExpRate on the testing dataset are 25.715% and 28.216%, respectively. Our result has reached the advanced level of accuracy under the same conditions. In future, we plan to improve the system by making good use of online ink trajectory information and try to complete a handwritten mathematical expression recognition software.

**Acknowledgements**

The authors want to thank Dr. Jianshu Zhang for insightful comments and suggestion. This work is supported by the National Key R&D Program of China (Grant No. 2016YFE0204200).